\documentclass[a4paper,conference]{IEEEtran}

\usepackage{graphicx}
\usepackage{amsmath,amssymb} 

\usepackage{algorithm}
\usepackage{amsfonts}

\usepackage{epsfig}
\usepackage{graphicx}
\usepackage{amsmath}
\usepackage{amssymb}
\usepackage{mathtools}
\usepackage{resizegather}
\usepackage{marvosym}
\usepackage{dsfont}
\usepackage{enumitem}
\usepackage{caption}
\usepackage{multirow}
\usepackage{amsthm}
\theoremstyle{plain}

\usepackage{caption}

\ifCLASSINFOpdf
\else
\fi
\begin{document}
%
\title{Energy-constrained Self-training for Unsupervised Domain Adaptation}

\author{\IEEEauthorblockN{Xiaofeng Liu$^{1,6\dag}$, Bo Hu$^{1,3\dag}$, Xiongchang Liu$^{4}$, Jun Lu$^{1*}$, Jane You$^5$, Lingsheng Kong$^2$}\\
\IEEEauthorblockA{$^1$ Beth Israel Deaconess Medical Center and Harvard Medical School, Boston, MA, USA\\$^2$ Changchun Institute of Optics, Fine Mechanics and Physics, Chinese Academy of Sciences, CAS, Changchun, China\\
$^3$ Illinois Institute of Technology, Chicago, IL, USA\\$^4$ Dept. of Information and Electrical Engineering, China University of Mining and Technology, China\\
$^5$ Dept. of Computing, Hong Kong Polytechnic University, Hung Hom, Hong Kong\\$^6$ Fanhan Tech. Inc., Suzhou, Jiangsu, China.\\
$\dag$ contribute equally. *Corresponding Author.}
}


%


\maketitle

\begin{abstract}
Unsupervised domain adaptation (UDA) aims to transfer the knowledge on a labeled source domain distribution to perform well on an unlabeled target domain. Recently, the deep self-training involves an iterative process of predicting on the target domain and then taking the confident predictions as hard pseudo-labels for retraining. However, the pseudo-labels are usually unreliable, and easily leading to deviated solutions with propagated errors. In this paper, we resort to the energy-based model and constrain the training of the unlabeled target sample with the energy function minimization objective. It can be applied as a simple additional regularization. In this framework, it is possible to gain the benefits of the energy-based model, while retaining strong discriminative performance following a plug-and-play fashion. We deliver extensive experiments on the most popular and large scale UDA benchmarks of image classification as well as semantic segmentation to demonstrate its generality and effectiveness.\vspace{+10pt}
\end{abstract}


%
\IEEEpeerreviewmaketitle

\begin{figure}[t]
\centering
\includegraphics[width=8cm]{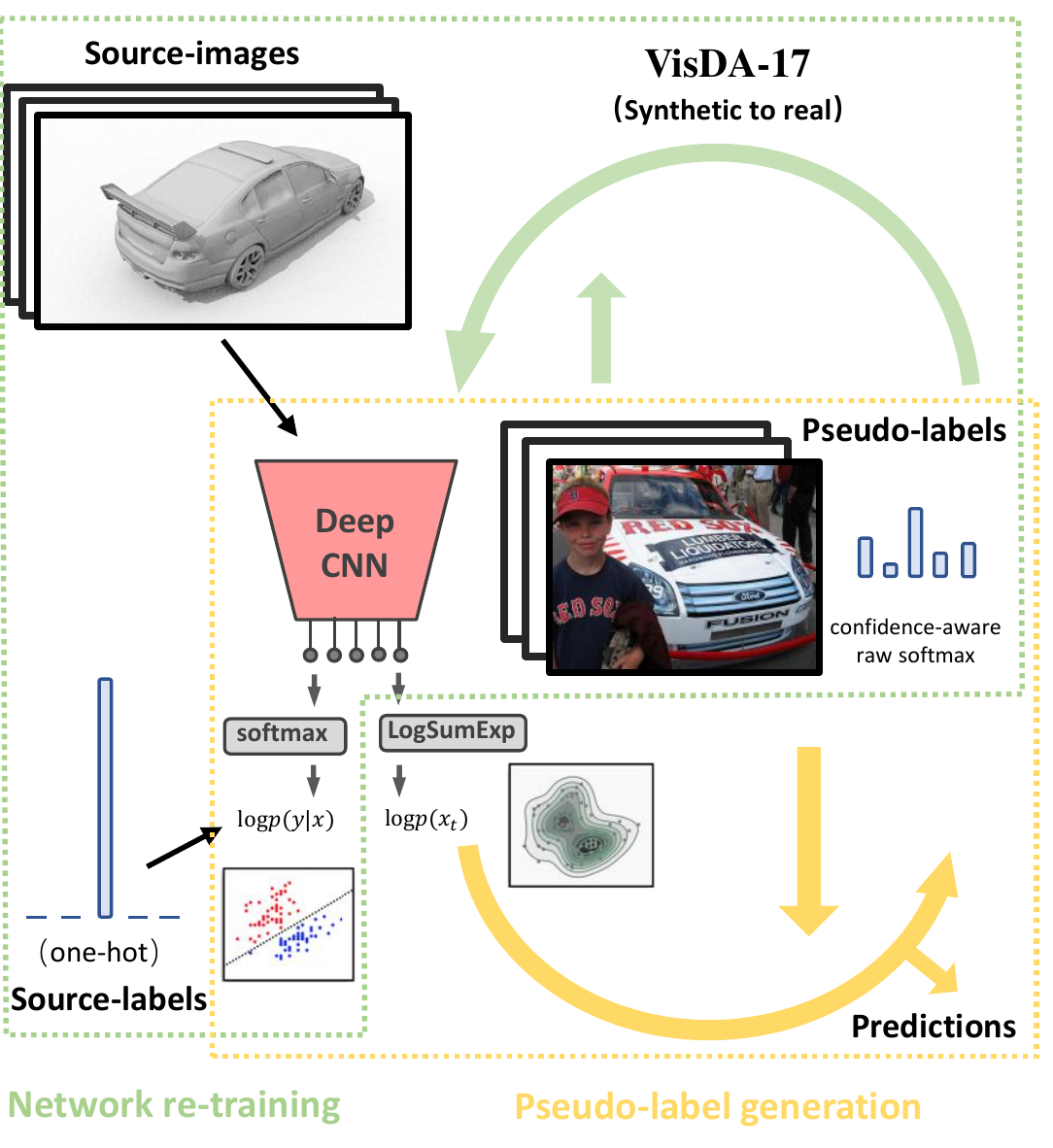}\\
\caption{The illustration of our Energy-constrained Self-training framework for UDA. Minimizing the pseudo label-irrelevant energy of $E_{\mathbf{w}}(\mathbf{x}_t)$ is introduced as additional objective for the target sample.}\label{fig:1} 
\end{figure}

\section{Introduction}

Deep neural networks are usually data-starved and rely on the $i.i.d$ assumption of training and testing data \cite{che2019deep,liu2020wasserstein,liu2020unimodal,liu2020importance,liu2020severity,he2020image,han2020wasserstein,liu2020auto3d}. However, in reality, the deployment target tasks are usually significantly diverse, and collecting labeled data in the target domain is expensive or even prohibitive. For instance, densely annotating a Cityscapes image on average takes about 90 minutes \cite{cordts2016cityscapes}, which severely hinders the generalization of an autonomous driving system in different cities.

Therefore, the unsupervised domain adaptation (UDA) seeks to transfer knowledge from one labeled source domain to another target domain with the aid of unlabeled target data \cite{kouw2018introduction}. Recently, one of the promising methods in UDA is the self-training \cite{lee2013pseudo,zou2019confidence}, which iteratively generates a set of one-hot pseudo-labels in the target domain, and then retrains network based on these pseudo-labels with target data. Previous research of deep self-training usually adopts the one-hot pseudo-label, and evidence that it is essentially an entropy minimization process \cite{lee2013pseudo} that pushing network output to be as sharp as hard pseudo-label.

However, the correctness of pseudo-labels cannot be guaranteed. Trusting all selected pseudo-labels as “ground truth” by encoding them as hard labels can lead to overconfident mistakes and propagated errors \cite{zou2019confidence}. Actually, the state-of-the-art accuracy of many UDA benchmarks are just around 50\%. Especially at the first few epochs, it is hard to produce reliable pseudo-label. Besides, the labels of natural images can also be highly ambiguous. Taking a sample image from VisDA17 \cite{peng2018visda} (see Fig. \ref{fig:1}) as an example, both person and car dominate significant portions of this image. Enforcing a model to be very confident in only one of the classes during training can also hurt the learning behavior \cite{bagherinezhad2018label}, particularly within the context of no ground truth label for the target samples. 

The manually defined pseudo label smoothing or entropy minimization are proposed to make the network more conservative \cite{zou2019confidence}. Essentially, our previous work \cite{zou2019confidence} is modifying the inaccurate pseudo label histogram distribution to be more smooth following a fixed smoothing operation for any data. For instance, revising the three-classes one-hot pseudo label $[1,0,0]$ or $[0,1,0]$ to $[0.9,0.05,0.05]$ or $[0.05,0.9,0.05]$. However, these constraints for pseudo-label can not be adaptively adjusted for different inputs or network parameters.

The aforementioned issues motivate us to introduce a regularization signal for the target sample that depends on the input and the network parameters at the present training iteration, and not related to the inaccurate pseudo label. 

Recent study \cite{grathwohl2019your} points out that a standard discriminative classifier can be essentially reinterpreted as an energy-based model (EBM) for the joint distribution $p(\mathbf{x},\mathbf{y})$ \cite{liu2018dependency,liu2019attention,liu2019permutation,liu2019dependency}. Both the energy-based model and supervised discriminative model can be benefited from simultaneously optimizing both $p(\mathbf{y}|\mathbf{x})$ with cross-entropy (CE) loss and optimizing ${\rm log}p(\mathbf{x})$ with EBM \cite{grathwohl2019your}. Here, we demonstrate that optimizing ${\rm log}p(\mathbf{x})$ can be even more promising on self-training-based UDA. Since we do not have reliable labels on target domain, and ${\rm log}p(\mathbf{x})$ can be an idea regularization signal, which is correlated with the input $\mathbf{x}$ and network parameter, and independent to pseudo label $\mathbf{y}$.

Therefore, we propose a simple and straightforward idea that configures the energy minimization of data point $\mathbf{x}$ (i.e., ${\rm log}p(\mathbf{x})$) as an additional regularization term. The target domain examples are optimized with pseudo-label CE loss and EBM objective. With the help of the energy-based model, our self-training is expected to be more controllable.

In this paper, we propose a novel and intuitive framework to incorporate the EBM into the self-training UDA as pseudo label-irrelevant adaptive regularization signal. We empirically validate the effectiveness and generality of the proposed method on multiple challenging benchmarks (classification and semantic segmentation) and achieve state-of-the-art performance.


\section{Related Works}

\noindent\textbf{Unsupervised domain adaptation (UDA)} with deep networks \cite{he2020classification,liu2019unimodal,liu2019conservative,liu2018ordinal,liu2018data} targets to learn domain invariant embeddings by minimizing the cross-domain difference of feature distributions with certain criteria \cite{liu2017line,liu2018joint}. Examples of these methods include maximum mean discrepancy (MMD), deep Correlation Alignment (CORAL), sliced Wasserstein discrepancy, adversarial learning at input-level, feature level, output space level, etc \cite{kouw2018introduction,liu2019feature}. Despite the underlying difference, there exists an interesting connection between some of these methods with conditional forms\footnote{E.g., class-wise adversarial learning, or discriminators taking network predictions as input.} and Self-training as they can be broadly considered as EM algorithms, and such conditional formulation has been widely proved to benefit the adaptation.\vspace{+5pt}

\noindent\textbf{Self-training} was initially investigated in semi-supervised learning \cite{triguero2015self}. A subtle difference between self-training with fixed feature input and deep self-training is that the latter involves the learning of deep embedding which renders greater flexibility towards domain alignment than classifier-level adaptation only. Recently, there have been multiple deep self-training/pseudo-label based methods that are proposed for UDA \cite{busto2018open,han2019unsupervised,Zou_2018_ECCV}.

Considering the noisy pseudo-label, our previous work \cite{zou2019confidence} proposes to construct a more conservative pseudo-label that smoothing the one-hot distribution or regularize it with the entropy. The solution proposes in this paper is orthogonal with \cite{zou2019confidence}, which resorts to the additional supervision signal of EBM that independent of pseudo-label. Compared with the manually defined label smoothing in \cite{zou2019confidence}, the energy-constraint can adaptively regularize the training w.r.t. the input and the present network parameters. We note that our EBM regularization can be simply added on state-of-the-art self-training methods following a plug-and-play fashion. \vspace{+5pt}

\noindent\textbf{Energy-Based Models (EBMs)} capture dependencies between variables by associating a scalar energy to each configuration of the variables \cite{lecun2006tutorial}. Recently, \cite{nijkamp2019learning,grathwohl2019your} approximated the expectation of the log-likelihood for a single example $\mathbf{x}$ with respect to network parameter using a sampler based on Stochastic Gradient Langevin Dynamics (SGLD) \cite{welling2011bayesian}.

Targeting on the combination of classifier and EBMs, \cite{xie2016theory,du2019implicit} reinterpret the logits to define a class-conditional EBM $p(\mathbf{x}|\mathbf{y})$, which require additional parameters to be learned to derive a classifier and an unconditional model. \cite{song2018learning} is similar as well but is trained using a GAN-like generator and is applied to different applications. Based on \cite{nijkamp2019learning}, \cite{grathwohl2019your} scales the training of EBMs to high-dimensional data with Contrastive Divergence and SGLD. 

Here, we regard EBMs as the pseudo label-irrelevant optimization constrains and explore its potentials on domain adaptation, especially UDA. This helps realize the potential of energy-based models on downstream supervised discriminative problems. Moreover, the thorough convergence analysis and theoretically connection with CEM are investigated.

\section{Methodology}

In the UDA setting, we can access to the labeled source samples $(\mathbf{x}_s,\mathbf{y}_s)$ from source domain $\{\mathbf{X}_S, \mathbf{Y}_S\}$, and target samples $\mathbf{x}_t$ from unlabeled target domain data $\mathbf{X}_T$. Any target label $\hat{\mathbf{y}}_t=(\hat{y}_t^{(1)},...,\hat{y}_t^{(K)})$ from $\hat{\mathbf{Y}}_T$ is unknown. $K$ is the total number of classes. The parametric network $f_{\mathbf{w}}:\mathbb{R}^D\rightarrow \mathbb{R}^K$ with the weights $\mathbf{w}$ is used to process the $D$-dim input sample. The output $K$ real-valued numbers known as logits and usually followed by a softmax normalization to produce ${p_\mathbf{w}}(k|\mathbf{x})={\rm exp}({f_{\mathbf{w}}}({\mathbf{x}})[k]) / {\sum_{k\in K}}{\rm exp}({f_{\mathbf{w}}}({\mathbf{x}})[k])$ as the classifier's softmax probability for class $k$. $f_{\mathbf{w}}({\mathbf{x}})[k]$ indicates the $k^{th}$ index of $f_{\mathbf{w}}({\mathbf{x}})$, i.e., the logit corresponding the the $k^{th}$ class label.

The recent study \cite{grathwohl2019your} reveals that a standard classifier can be interpreted as an energy based model of joint distribution $p_{\mathbf{w}}(\mathbf{x},\mathbf{y})$, and optimizing the likelihood ${\rm log}p_{\mathbf{w}}(\mathbf{x},\mathbf{y})={\rm log}p_{\mathbf{w}}(\mathbf{x})+{\rm log}p_{\mathbf{w}}(\mathbf{y}|\mathbf{x})$ can be helpful for both the discrimination and generation task. Specifically, optimizing $p(\mathbf{y}|\mathbf{x})$ is simply achieved by using the standard cross-entropy loss as conventional classification. The additional optimization objective ${\rm log}p_{\mathbf{w}}(\mathbf{x})$ has been proven and evidenced that can improve the confidence calibration and robustness for conventional classification task \cite{grathwohl2019your}.  

Considering the target samples do not have ground truth label, the self-training methods \cite{zou2019confidence} utilize the inaccurate pseudo label to calculate the cross-entropy loss. Therefore, optimizing ${\rm log}p_{\mathbf{w}}(\mathbf{x})$ can potentially be more helpful for UDA setting. Actually, ${\rm log}p_{\mathbf{w}}(\mathbf{x})$ is adaptive w.r.t. the input $\mathbf{x}$ and network parameter $\mathbf{w}$, and irrelevant to the inaccurate pseudo label, which can be an ideal regularizer of self-training based UDA. 

However, how to modeling ${\rm log}p_{\mathbf{w}}(\mathbf{x})$ can be a challenging task. In EBM, the probability density $p_{\mathbf{w}}(\mathbf{x})$ for $\mathbf{x}\in\mathbb{R}^D$ can be expressed as $p_{\mathbf{w}}(\mathbf{x})={\rm exp}(-E_{\mathbf{w}}(\mathbf{x}))/Z(\mathbf{w})$. The energy based function $E_{\mathbf{w}}(\mathbf{x}):\mathbb{R}^D\rightarrow\mathbb{R}$ maps each point of an input space to a single scalar, which is called “energy”. One can parameterize an EBM using any function that takes $\mathbf{x}$ as the input and returns a scalar. $Z(\mathbf{w})=\int_{\mathbf{x}}{\rm exp}(-E_{\mathbf{w}}(\mathbf{x}))$ is the normalizing constant (with respect to $\mathbf{x}$) also known as the partition function. Usually, the normalized densities $p_{\mathbf{w}}(\mathbf{x})$ are intractable, since we cannot reliably estimate $Z(\mathbf{w})$ for the most choice of $E_{\mathbf{w}}(\mathbf{x})$. Usually we rely on the sophisticate Markov Chain Monte Carlo sampler to train EBMs.

Considering $\frac{\partial{\rm log}p_{\mathbf{w}}(\mathbf{x})}{\partial{\mathbf{w}}}$ can be approximated with $-\frac{\partial E_{\mathbf{w}}(\mathbf{x})}{\partial{\mathbf{w}}}$ \cite{grathwohl2019your}, it is possible to modeling the energy function $E_{\mathbf{w}}(\mathbf{x})$ instead of ${\rm log}p_{\mathbf{w}}(\mathbf{x})$. Following \cite{grathwohl2019your}, we can define an EBM of the joint distribution $p_{\mathbf{w}}(\mathbf{x},\mathbf{y})={\rm exp}(f_{\mathbf{w}}({\mathbf{x}})[k])/Z({\mathbf{w}})$, by defining $E_{\mathbf{w}}(\mathbf{x},\mathbf{y})=-f_{\mathbf{w}}({\mathbf{x}})[k]$. By marginalizing out $\mathbf{y}$, we have $p_{\mathbf{w}}(\mathbf{x})=\frac{\sum_k{\rm exp}(f_{\mathbf{w}}({\mathbf{x}})[k])}{Z({\mathbf{w}})}$ \cite{grathwohl2019your}. Considering $p_{\mathbf{w}}(\mathbf{x})={\rm exp}(-E_{\mathbf{w}}(\mathbf{x}))/Z(\mathbf{w})$, the energy function of $\mathbf{x}$ can be
\begin{align}\label{energy} E_{\mathbf{w}}(\mathbf{x})=-{\rm log}\sum_k{\rm exp}{(f_{\mathbf{w}}({\mathbf{x}})[k])}
\end{align}

In this setting, $p_{\mathbf{w}}(\mathbf{x}|\mathbf{y})=\frac{p_{\mathbf{w}}(\mathbf{x},\mathbf{y})}{p_{\mathbf{w}}(\mathbf{x})}=\frac{{\rm exp}(f_{\mathbf{w}}({\mathbf{x}})[k])/Z({\mathbf{w}})}{{\sum_k{\rm exp}(f_{\mathbf{w}}({\mathbf{x}})[k])}/{Z({\mathbf{w}})}}$. The normalization constant $Z({\mathbf{w}})$ will be canceled out and yielding the standard softmax function, which bridges the EMB and conventional classifiers.

Therefore, we can simultaneously optimize $p(\mathbf{y}|\mathbf{x})$ with standard cross-entropy loss, and optimize ${\rm log}p_{\mathbf{w}}(\mathbf{x})$ with Stochastic Gradient Langevin Dynamics (SGLD), where gradients are taken with respect to $-{\rm log}\sum_k{\rm exp}{(f_{\mathbf{w}}({\mathbf{x}})[k])}$ \cite{grathwohl2019your}.


\begin{table*}[t]
\centering
\resizebox{\linewidth}{!}{
\centering
\begin{tabular}{c|cccccccccccc|c}
\hline
Method         &  Aero & Bike   & Bus & Car & Horse & Knife & Motor   & Person   & Plant & Skateboard & Train  & Truck   & Mean \\ \hline\hline
Source-Res101 \cite{gholami2019taskdiscriminative} & 55.1 & 53.3 & 61.9 & 59.1 & 80.6 & 17.9 & 79.7 & 31.2 & 81.0 & 26.5 & 73.5 & 8.5 & 52.4 \\\hline

MMD \cite{long2015learning} & 87.1 & 63.0 & 76.5 & 42.0 & 90.3 & 42.9 & 85.9 & 53.1 & 49.7 & 36.3 & 85.8 & 20.7 & 61.1 \\
DANN \cite{ganin2016domain} & 81.9 & 77.7 & 82.8 & 44.3 & 81.2 & 29.5 & 65.1 & 28.6 & 51.9 & 54.6 & 82.8 & 7.8 & 57.4 \\ 
ENT \cite{grandvalet2005semi} & 80.3 & 75.5 & 75.8 & 48.3 & 77.9 & 27.3 & 69.7 & 40.2 & 46.5 & 46.6 & 79.3 & 16.0 & 57.0 \\
MCD \cite{saito2017maximum} & 87.0 & 60.9 & \textbf{83.7} & 64.0 & 88.9 & 79.6 & 84.7 &  {76.9} &  {88.6} & 40.3 & 83.0 & 25.8 & 71.9 \\
ADR \cite{saito2018adversarial} & 87.8 & 79.5 & \textbf{83.7} & 65.3 & \textbf{92.3} & 61.8 &  {88.9} & 73.2 & 87.8 & 60.0 & \textbf{85.5} & {32.3} & 74.8 \\  
DEV \cite{you2019toward} & 81.83& 53.48& 82.95& 71.62& 89.16 &72.03& \textbf{89.36}& 75.73& \textbf{97.02} & 55.48 &71.19 &29.17& 72.42\\
TDDA \cite{gholami2019taskdiscriminative} & 88.2 &78.5& 79.7& 71.1 &90.0 &81.6& 84.9 &72.3 &92.0& 52.6& 82.9& 18.4 &74.03\\\hline

CBST \cite{Zou_2018_ECCV} & 87.1$\pm$1.2 & 79.5$\pm$2.3 & 58.3$\pm$2.6 & 50.4$\pm$3.9 & 82.8$\pm$2.1 & 73.7$\pm$7.2 & 80.9$\pm$2.6 & 71.8$\pm$3.1 & 81.6$\pm$3.2 & 88.4$\pm$3.3 & 75.2$\pm$1.2 & 68.4$\pm$3.4 & 74.8$\pm$0.5 \\

\hline

\textbf{CBST+$R_{EBM}$} & 87.9$\pm$1.6 & 79.6$\pm$1.5 & 68.5$\pm$1.3 & 68.6$\pm$1.9 & 83.2$\pm$1.2 & {78.4$\pm$1.9} & 83.5$\pm$1.5 & 72.2$\pm$1.5 & 82.2$\pm$1.6 & 84.3$\pm$1.5 & 80.9$\pm$1.4 & 67.5$\pm$1.3 & 77.0$\pm$0.6 \\
 
\hline

CRST\cite{zou2019confidence} & 89.2$\pm$1.6 & {79.6$\pm$4.6} & 64.2$\pm$4.0 & 57.8$\pm$3.4 & 87.8$\pm$1.9 & 79.6$\pm$8.5 & 85.6$\pm$2.6 & 75.9$\pm$4.2 & 86.5$\pm$2.2 & 85.1$\pm$2.4 & 77.7$\pm$2.2 & 68.5$\pm$0.9 & {78.1$\pm$0.7} \\

\hline

\textbf{CRST+$R_{EBM}$} & \textbf{90.3}$\pm$1.5 & \textbf{82.6$\pm$1.2} & 72.4$\pm$1.5 & \textbf{71.7$\pm$1.8} & 87.6$\pm$1.8& \textbf{81.8$\pm$1.9} & 85.4$\pm$1.5 & \textbf{80.8$\pm$1.5} & 87.1$\pm$1.6 & \textbf{89.9$\pm$1.5} & 83.6$\pm$1.6 & \textbf{71.5$\pm$1.3 }& \textbf{80.2$\pm$0.5}\\

\hline
\hline

SimNet* \cite{pinheiro2018unsupervised} & \textbf{94.3} & 82.3 & 73.5 & 47.2 & 87.9 & 49.2 & 75.1 & 79.7 & 85.3 & 68.5 & 81.1 & 50.3 & 72.9 \\

GTA* \cite{sankaranarayanan2018generate}  
& - & - & - & - & - & - & - & - & - & - & - & - & 77.1\\\hline
\textbf{CRST+$R_{EBM}$* }& {93.2$\pm$1.3} & \textbf{85.8$\pm$1.2} &  {73.7$\pm$1.0 }& \textbf{74.3$\pm$1.5 }& {89.5$\pm$0.6} & \textbf{87.6$\pm$1.6} &  {88.2$\pm$1.6 }& \textbf{82.2$\pm$1.6 }&  {90.9$\pm$1.3} & \textbf{91.6$\pm$1.8} & {85.1$\pm$1.4} & \textbf{79.9$\pm$1.4} & \textbf{82.8$\pm$0.5 }\\
\hline

\end{tabular}%
}
\caption{Experimental results for VisDA17-val setting. We use ResNet101 as backbone except SimNet and GTA.*ResNet152 backbone.}
\label{table:visda17}
\end{table*}

\subsection{EBM as regularization for UDA}

Self-training for UDA is an iterative loss minimization framework \cite{zou2019confidence}, which regards pseudo-labels as learnable latent variables. The optimization objective for unlabeled target example is the cross-entropy with one-hot \cite{Zou_2018_ECCV} or smoothed \cite{zou2019confidence} pseudo-label. Therefore, a straightforward solution for adapting EBM to UDA is to incorporate Eq. \ref{energy} as a regularization term. For the target sample, it not only needs to achieve a good prediction of pseudo-label, but also minimize the energy of $E_{\mathbf{w}}(\mathbf{x}_t)$. Therefore, our energy-constraint is essentially depended on $\mathbf{x}$ and $\mathbf{w}$, which is more flexible than pre-defined label smoothing or entropy regularization \cite{zou2019confidence}. Considering that the pseudo-label is usually noisy, the latter objective is expected to be even more important than the setting of supervised learning.

Following the formulation in our CRST \cite{zou2019confidence}, the self-training with EBM regularization (R-EBM) for target sample, i.e., $E_{\mathbf{w}}(\mathbf{x}_t)$, can be formulated as \begin{align}\label{cbst}
&\underset{\mathbf{w},{{{\hat{\mathbf{Y}}}}_{T}}}{\mathop{\min }}\,\mathcal{L}_{R-EBM}(\mathbf{w}, {\hat{\mathbf{Y}}}) = -\sum\limits_{{{s}}\in {{S}}}{\sum\limits_{k=1}^{K}{y_{s}^{(k)}}\log p_\mathbf{w}(k|{\mathbf{x}_{s}})} \nonumber \\
&-\sum\limits_{{{t}}\in {{T}}}\{{\sum\limits_{k=1}^{K}{[\hat{y}_{t}^{(k)}}\log p_\mathbf{w}(k|{\mathbf{x}_{t}})}-\hat{y}_{t}^{(k)}{{\log\lambda }_{k}}]-\alpha E_{\mathbf{w}}(\mathbf{x}_t)\}  \nonumber \\
&s.t.~{{{\hat{\mathbf{y}}}}_{t}}\in \Delta^{K-1}\cup \{\mathbf{0}\},\forall t 
\end{align} For each class $k$, $\lambda_{k}$ is determined by the confidence value selecting the most confident $p$ portion of class $k$ predictions in the entire target set \cite{Zou_2018_ECCV}. If a sample's predication is relatively confident with $p_\mathbf{w}(k^*|\mathbf{x}_t) > \lambda_{k^*}$, it is selected and labeled as class $k^* = \text{argmax}_{k}\{\frac {p_\mathbf{w}(k|{\mathbf{x}_{t}})}{\lambda_k}\}$. The less confident ones with $p_\mathbf{w}(k^*|\mathbf{x}_t) \leq \lambda_{k^*}$ are not selected. The same class-balanced $\lambda_{k}$ strategy introduced in \cite{Zou_2018_ECCV} is adopted for all self-training methods in this work. 

The feasible set is the union of $\{\mathbf{0}\}$ and a probability simplex $\Delta^{K-1}$  \cite{Zou_2018_ECCV}. $\alpha$ is a balancing hyper-parameter of the regularization term, which does not directly relate to the label $\hat{y}_{t}$. The self-training can be solved by an alternating optimization scheme.\vspace{+5pt}

\noindent \textbf{Step 1) Pseudo-label generation} \label{_a)} ~ Fix $\mathbf{w}$ and solve: \begin{align}\label{cbst_a}
& \underset{{{{\hat{\mathbf{Y}}}}_{T}}}{\mathop{\min }}  -\sum\limits_{{{t}}\in {{T}}}\{{\sum\limits_{k=1}^{K}{\hat{y}_{t}^{(k)}}[\log {p_\mathbf{w}(k|{\mathbf{x}_{t}})}-\log{{{\lambda }_{k}}}]}-\alpha E_{\mathbf{w}}(\mathbf{x}_t) \} \nonumber\\
& s.t.\text{ }{{{\hat{y}}}_{t}}\in\Delta^{K-1}\cup \{\mathbf{0}\},\forall t \end{align} 

For solving step \textbf{1)}, there is a global optimizer for arbitrary $\hat{\mathbf{y}}_t=(\hat{y}_t^{(1)},...,\hat{y}_t^{(K)})$ as \cite{zou2019confidence}:
\begin{equation}
\hat{y}_{t}^{(k)*}=\left\{
\begin{aligned}
1, &~\text{if}~k={~}^{\rm argmax}_{~~~~k}~{\frac {p_\mathbf{w}(k|{\mathbf{x}_{t}})}{\lambda_k}}\\
& ~~~~~ \text{and} ~~~ p_\mathbf{w}(k|\mathbf{x}_t)>\lambda_k\\
0, &~\mathrm{otherwise}
\end{aligned}
\right.\label{cbst_a_solver}
\end{equation}

Despite a long period of little development, there has been recent work \cite{du2019implicit,nijkamp2019learning} using the sampler based on SGLD to train the large-scale EBMs on high-dimensional data, parameterized by deep neural networks.

Given that the goal of our work is to incorporate EBM training into the standard classification setting, the classification part is the same as \cite{zou2019confidence}, and we only change the regularizer. Therefore, we can follow \cite{grathwohl2019your} to train the network with both cross-entropy loss and SGLD to ensure this distribution is being optimized with an unbiased objective. Similar to \cite{du2019implicit} we also adopt the contrastive divergence to estimate the expectation of derivative of the log-likelihood for a single example $\mathbf{x}$ with respect to $\mathbf{w}$. Since it gives an order of magnitude savings in computation compared to seeding new chains at each iteration as in \cite{nijkamp2019learning}.\vspace{+5pt}

\noindent \textbf{Step 2) Network retraining} \label{b)} ~ Fix $\hat{\mathbf{Y}}_T$ and minimize \begin{align}\tiny\label{cbst_b}
-\sum\limits_{{{s}}\in {{S}}}{\sum\limits_{k=1}^{K}{y_{s}^{(k)}}\log p_\mathbf{w}(k|{\mathbf{x}_{s}})}-\sum\limits_{{{t}}\in {{T}}}{\sum\limits_{k=1}^{K}{\hat{y}_{t}^{(k)}}\log p_\mathbf{w}(k|{\mathbf{x}_{t}})} 
\end{align}

\noindent w.r.t. $\mathbf{w}$. Carrying out step \textbf{1)} and \textbf{2)} for one time is defined as one round in self-training.

\section{Experiments}

\begin{table*}[t]
\centering
\resizebox{\linewidth}{!}{
\centering
\begin{tabular}{c|c|ccccccccccccccccccc|c}
\hline
Method  & Base Net  & Road & SW   & Build & Wall & Fence & Pole & TL   & TS   & Veg. & Terrain & Sky  & PR   & Rider & Car  & Truck & Bus  & Train & Motor & Bike & mIoU \\ \hline\hline
Source     & DRN26 & 42.7 & 26.3 & 51.7  & 5.5  & 6.8   & 13.8 & 23.6 & 6.9  & 75.5 & 11.5    & 36.8 & 49.3 & 0.9   & 46.7 & 3.4   & 5.0  & 0.0   & 5.0   & 1.4  & 21.7 \\
CyCADA \cite{hoffman2018cycada}         &   & 79.1 & 33.1 & 77.9  & 23.4 & 17.3  & 32.1 & 33.3 & 31.8 & 81.5 & 26.7 & 69.0 & 62.8 & 14.7  & 74.5 & 20.9  & 25.6 & 6.9   & 18.8  & 20.4 & 39.5 \\ \hline
Source     & DRN105          &  36.4 & 14.2 & 67.4 & 16.4 & 12.0 & 20.1 & 8.7 & 0.7 & 69.8 & 13.3 & 56.9 & 37.0 & 0.4 & 53.6 & 10.6 & 3.2 & 0.2 & 0.9 & 0.0 & 22.2 \\
MCD \cite{saito2017maximum}   &   &90.3 & 31.0 & 78.5 & 19.7 & 17.3 & 28.6 & 30.9 & 16.1 & 83.7 & 30.0 & 69.1 & 58.5 & 19.6 & 81.5 & 23.8 & 30.0 & 5.7 & 25.7 & 14.3 & 39.7 \\ \hline
Source & PSPNet & 69.9 & 22.3 & 75.6 & 15.8 & 20.1 & 18.8 & 28.2 & 17.1 & 75.6 & 8.00 & 73.5 & 55.0 & 2.9 & 66.9 & \textbf{34.4} & 30.8 & 0.0 & 18.4 & 0.0 & 33.3 \\
DCAN \cite{Wu_2018_ECCV} & & 85.0 & 30.8 & 81.3 & 25.8 & 21.2 & 22.2 & 25.4 & 26.6 & 83.4 & {36.7} & 76.2 & 58.9 & 24.9 & 80.7 & 29.5 & 42.9 & 2.50 & 26.9 & 11.6 & 41.7 \\ \hline

Source     & DeepLabv2          &  75.8 & 16.8 & 77.2 & 12.5 & 21.0 & 25.5 & 30.1 & 20.1 & 81.3 & 24.6 & 70.3 & 53.8 & 26.4 & 49.9 & 17.2 & 25.9 & 6.5 & 25.3 & 36.0 & 36.6\\
AdaptSegNet \cite{Tsai_adaptseg_2018}   & & 86.5 & 36.0 & 79.9& 23.4 &  {23.3} & 23.9 & 35.2 & 14.8 & 83.4 & 33.3 & 75.6 & 58.5 & 27.6 & 73.7 & 32.5 & 35.4 & 3.9 & 30.1 & 28.1 & 42.4 \\ \hline

AdvEnt \cite{vu2019advent}   & DeepLabv2 & 89.4 & 33.1 & 81.0 & 26.6 & 26.8 & 27.2 & 33.5 & 24.7 & 83.9 & 36.7 & 78.8 & 58.7 & 30.5 & 84.8 & 38.5 & 44.5 & 1.7 & 31.6 & 32.4 & 45.5 \\ \hline

Source     & DeepLabv2          &  - & - & - & - & - & - & - & - & -& - & - & - & - & - & - & - & - & - & - & 29.2\\
FCAN \cite{zhang2018fully}   &  &  - & - & - & - & - & - & - & - & -& - & - & - & - & - & - & - & - & - & - &  {46.6} \\ \hline

Source&DeepLabv2 &75.8 &16.8 &77.2 &12.5 &21.0&25.5 &30.1 &20.1 &81.3 &24.6 &70.3& 53.8& 26.4 &49.9 &17.2 &25.9& 6.5& 25.3 &36.0 &36.6\\

DPR \cite{tsai2019domain} & &92.3 &51.9 &\textbf{82.1} &29.2& 25.1& 24.5& 33.8& \textbf{33.0} &82.4& 32.8& \textbf{82.2}& 58.6& 27.2& 84.3 &33.4& \textbf{46.3} &2.2 &29.5 &32.3& 46.5\\\hline

Source&DeepLabv2 &73.8& 16.0& 66.3& 12.8 &22.3& 29.0& 30.3 &10.2 &77.7 &19.0 &50.8 &55.2& 20.4& 73.6& 28.3 &25.6 &0.1& 27.5 &12.1 &34.2\\

PyCDA \cite{lian2019constructing} && 90.5&  36.3&  84.4&  32.4 & \textbf{28.7} & 34.6&  36.4 & 31.5&  86.8&  37.9&  78.5 & 62.3 & 21.5&  \textbf{85.6}&  27.9 & 34.8&  18.0 & 22.9&  \textbf{49.3} & 47.4\\\hline
\hline

Source &DeepLabv2 & 71.3 & 19.2 & 69.1 & 18.4 & 10.0 & 35.7 & 27.3 &  6.8 & 79.6 & 24.8 & 72.1 & 57.6 & 19.5 & 55.5 & 15.5 & 15.1 & 11.7 & 21.1 & 12.0 & 33.8 \\

CBST \cite{Zou_2018_ECCV}      &                   & 89.9 &  55.0 &  79.9 &  29.5 &  20.6 &  37.8 &  32.9 &  13.9 &  84.0 &  31.2 &  75.5 &  60.2 &  27.1 &  81.8 &  29.7 &  40.5 &  7.62 &  28.7 &  41.4 & 45.6 \\

CBST+$R_{EBM}$      &   & 91.1 & 53.9 & 80.6 & 31.6 & 21.0 & 40.4 & 35.0 & 19.8 & {86.8} & 35.9 & 76.4 &\textbf{63.3}& 31.4 & 83.0 & 22.5 & 38.6 & 24.2 & 32.2 & 39.4 & 47.8 \\\hline

Source &DeepLabv2 & 71.3 & 19.2 & 69.1 & 18.4 & 10.0 & 35.7 & 27.3 &  6.8 & 79.6 & 24.8 & 72.1 & 57.6 & 19.5 & 55.5 & 15.5 & 15.1 & 11.7 & 21.1 & 12.0 & 33.8 \\

CRST \cite{zou2019confidence}    &                   & 89.0 & 51.2 & 79.4 &  \textbf{31.7} & 19.1 & 38.5 & 34.1 & 20.4 & 84.7 & 35.4 & {76.8} & 61.3 & {30.2} & 80.7 & 27.4 & 39.4 & 10.2 & 32.2 & {43.3} & {46.6} \\

CRST+$R_{EBM}$    &         & \textbf{92.5} & \textbf{56.6} & 80.9 & 26.2 & 20.5 & \textbf{40.5} & \textbf{35.3} & 24.4 & \textbf{86.9} & \textbf{37.3} & 77.5 & 63.4 & \textbf{30.5} & 81.3 & 28.8 & 39.2 & \textbf{24.6} & \textbf{33.5} & 41.3 & \textbf{48.5}\\\hline
\end{tabular}
}
\caption{Experimental results for GTA5 to Cityscapes.}
\label{table:gtacity}
\end{table*}

\begin{figure*}[t]
\centering
\includegraphics[width=18cm]{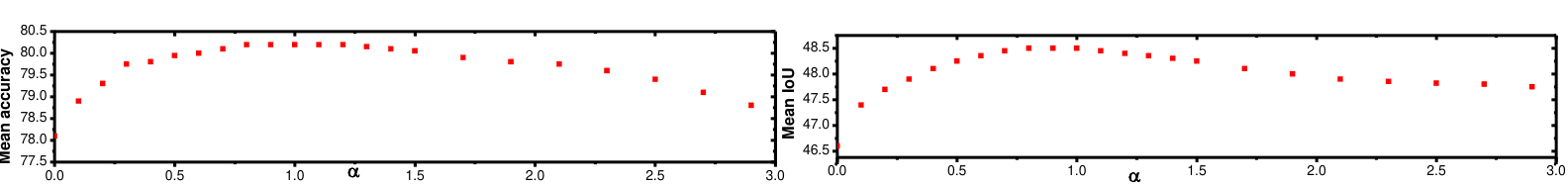}\\
\caption{Sensitive analysis of hyper-parameter $\alpha$ in VisDA17 (left) and CTA52Sityscapes (right) with CRST+$R_{EBM}$.}\label{fig:3} \vspace{+10pt}
\end{figure*}

In this section, we provide comprehensive evaluations of the proposed EBM objective both on image classification and semantic segmentation UDA tasks. We implement our methods using the PyTorch toolbox \cite{paszke2017automatic}.

\subsection{Domain Adaptation for Image Classification}


The VisDA17 \cite{peng2018visda} benchmark is a 12-class UDA classification problem. We follow the standard protocol in \cite{zou2019confidence,sankaranarayanan2018generate} where the source domain is the training set including $152,409$ synthetic 2D images and the target domain is the validation set including $55,400$ real images from COCO dataset.

To make a fair comparison with other methods, we use the same backbone network (e.g., ResNet101 or ResNet152) for VisDA17. Both networks are pre-trained in ImageNet as previous works and fine-tuned in source domain by SGD with a fixed learning rate $1\times10^{-3}$, weight decay $5\times10^{-4}$, momentum $0.9$ and batch size $32$. As shown in Fig. \ref{fig:3}, the performance is not sensitive to $\alpha$ when $\alpha\in[0.8,1.1]$, we simply choose $\alpha=1$ in all settings.

We present the results on VisDA 17 in Table \ref{table:visda17} in terms of per-class accuracy and mean accuracy. For each proposed approach, we independently run 5 times and the average and standard deviation of the accuracy metrics are reported.

Class-balanced self-training (CBST) and Conservative regularized self-training (CRST) \cite{zou2019confidence} are the state-of-the-art self-training UDA methods. The CRST+$R_{EBM}$ denote the CRST with EBM regularization.

Benefited by the additional EMB objective, CBST/CRST +$R_{EBM}$ can outperform CBST/CRST significantly, and outperforms the recent UDA methods other than self-training by a large margin. We note that the adversarial training can further boost the performance of self-training \cite{zou2019confidence}. Our energy constraint can adaptively regularize the training w.r.t. the input $\mathbf{x}$ and the present network parameters, which is more flexible than the manually pre-defined label smoothing \cite{zou2019confidence}.

More appealingly, CRST+$R_{EBM}$ achieves the better performance than the recent adversarial training \cite{gholami2019taskdiscriminative}, dropout \cite{saito2017adversarial} and moment matching methods, which revokes the potential of self-training in UDA.

The more powerful backbones have also been applied and show better results \cite{pinheiro2018unsupervised,sankaranarayanan2018generate}. The EBM objective with ResNet152 backbone outperforms the other state-of-the-arts \cite{pinheiro2018unsupervised,sankaranarayanan2018generate}. It also demonstrate the flexibility of $R_{EBM}$ for different backbones.




Actually, our $R_{EBM}$ can be orthogonal with the recent progress of self-training based UDA. CBST \cite{Zou_2018_ECCV} is a pioneer of vanilla adversarial UDA, and its performance on VisDA17 is reported on \cite{zou2019confidence}. The label smoothing or the entropy regularization used in CRST \cite{zou2019confidence} or the vanilla self-training UDA \cite{Zou_2018_ECCV} can be simply add-on our $R_{EBM}$ to further improve the performance. CRST+$R_{EBM}$ obtains better or competitive performances in all settings, even compared with a generative pixel-level domain adaptation method GTA, which is a very complex algorithm in both architecture and objectives.

\subsection{UDA for Semantic Segmentation}

The semantic segmentation is essentially making the pixel-wise classification. We consider the challenging segmentation adaptation settings in GTA5 \cite{richter2016playing} to Cityscapes (19 classes are shared). GTA5 dataset includes 24,966 annotated images with size 1,052$\times$1,914, which rendered by the GTA5 game engine. Following the standard protocols \cite{hoffman2018cycada,Tsai_adaptseg_2018}, we use the full set of GTA5 and adapt the model to the Cityscapes train set with $2,975$ images. In testing, we evaluate on the Cityscapes validation set with $500$ images. 

To make a fair comparison with other methods, we use the ResNet101 as the backbone network as \cite{Zou_2018_ECCV,zou2019confidence}. Noticing that in PSPNET \cite{zhao2017pyramid}, Wide ResNet38 is a stronger basenet than ResNet101. The basenet is pre-trained in ImageNet and fine-tuned in source domain by SGD with learning rate $2.5\times10^{-4}$, weight decay $5\times10^{-4}$, momentum $0.9$, batch size $2$, patch size $512\times1024$ and data augmentation of multi-scale training ($0.5 - 1.5$) and horizontal flipping. All results on this dataset in the main paper are unified to report the mIoU of the models at the end of the $6^{th}$ epoch.

We compare CBST/CRST+$R_{EBM}$ with the other methods in Table \ref{table:gtacity}. Based on the previous self-training UDA methods CBST or CRST, our additional EBM objective outperforms CBST or CRST by about 2\% w.r.t. the mean IoUs. 

We achieve a new state-of-the-art, even compared with the generative pixel-level domain adaptation method GTA \cite{zhang2018fully}, which is a relatively complex algorithm in both architecture and objectives. 

Introducing the EBM objective to self-training methods can significantly improve the segmentation performance. Consistently with the classification, $R_{EBM}$ can be better or comparable with adversarial learning, indicating the self-training can still be a powerful methodology in UDA. Note that moment matching-based methods are usually not well scaleable to segmentation.

\section{Conclusions}

In this paper, we propose to introduce the energy based model's objective into the self-training unsupervised domain adaptation. Considering the lack of target label, we resort to the pseudo-label which is usually noisy. The EBM optimization objective provides additional signal that is independent of pseudo-label. It can be even more promising in UDA setting than supervised learning. It can be a regularization term. Our solution is orthogonal with the recent progress of self-training and can be added on in a plug-and-play manner without introducing large computation and the change of network structure. Extensive experiments on both UDA classification and semantic segmentation evidenced its effectiveness and generality. The more advanced EBM training and self-training methods \cite{han2019unsupervised} can also be adopted to further improve the results.

\section{Acknowledgements}

This work was supported by the Jangsu Youth Programme [BK20200238], National Natural Science Foundation of China, Younth Programme [grant number 61705221], NIH [NS061841, NS095986], Fanhan Technology, and Hong Kong Government General Research Fund GRF (Ref. No.152202/14E) are greatly appreciated.

\bibliographystyle{IEEEtran}
\bibliography{IEEEabrv.bib}

\end{document}